\documentclass[runningheads]{llncs}
\usepackage[T1]{fontenc}
\usepackage{amsmath}
\usepackage{amssymb}
\usepackage{booktabs}
\usepackage{multirow} 
\usepackage{pifont}
\usepackage{hyperref}
\usepackage[table]{xcolor}
\newcommand{\xmark}{\ding{55}}
\usepackage{graphicx}
\newcommand{\RNum}[1]{\uppercase\expandafter{\romannumeral #1\relax}}
\usepackage{wrapfig}
\usepackage{enumitem}
\begin{document}
\title{VerSe: Integrating Multiple Queries as Prompts for Versatile Cardiac MRI Segmentation}
\titlerunning{VerSe: Versatile Cardiac MRI Segmentation}
%
\author{Bangwei Guo\inst{1} \and
Meng Ye\inst{1} \and
Yunhe Gao\inst{1} \and
Bingyu Xin\inst{1} \and
Leon Axel\inst{2} \and
Dimitris Metaxas\inst{1}}
\authorrunning{F. Author et al.}
%
\institute{Rutgers University \\
\and
New York University School of Medicine
}
\maketitle              
\begin{abstract}
Despite the advances in learning-based image segmentation approach, the accurate segmentation of cardiac structures from magnetic resonance imaging (MRI) remains a critical challenge. While existing automatic segmentation methods have shown promise, they still require extensive manual corrections of the segmentation results by human experts, particularly in complex regions such as the basal and apical parts of the heart. Recent efforts have been made on developing interactive image segmentation methods that enable human-in-the-loop learning. However, they are semi-automatic and inefficient, due to their reliance on click-based prompts, especially for 3D cardiac MRI volumes. To address these limitations, we propose \texttt{VerSe}, a \texttt{Ver}satile \texttt{Se}gmentation framework to unify automatic and interactive segmentation through mutiple queries. Our key innovation lies in the joint learning of object and click queries as prompts for a shared segmentation backbone. \texttt{VerSe} supports both fully automatic segmentation, through object queries, and interactive mask refinement, by providing click queries when needed. With the proposed integrated prompting scheme, \texttt{VerSe} demonstrates significant improvement in performance and efficiency over existing methods, on both cardiac MRI and out-of-distribution medical imaging datasets. The code is available at \href{https://github.com/bangwayne/Verse}{https://github.com/bangwayne/Verse}.

\end{abstract}
\section{Introduction}
\label{sec:intro}
Cardiac magnetic resonance imaging (MRI) can provide comprehensive information for heart disease diagnosis and treatment~\cite{pennell2024cardiovascular}, serving as the gold standard for various clinical applications. 
For example, cardiac cine MRI~\cite{bernard2018deep} enables precise evaluation of cardiac function, while
cardiac late gadolinium enhancement (LGE) MRI~\cite{romero2022cmrsegtools} excels in detecting myocardium infarction (MI) and assessing tissue viability. 
Despite these advantages, the widespread clinical adoption of cardiac MRI lags behind echo cardiography and computational tomography (CT), in large part due to the challenges in image post-processing,~\textit{e.g.}, the accurate segmentation of anatomical structures and lesions. 

Deep learning-based methods have significantly improved automatic cardiac MRI segmentation. U-Net~\cite{ronneberger2015u} and its variants~\cite{gao2021utnet,isensee2021nnu,milletari2016v} remain the most widely used architecture and perform well on both 2D and 3D images, but struggle with intricate object boundaries and complex structures~\cite{bernard2018acdc}. Vision Transformer (ViT)-based models~\cite{chen2021transunet,dosovitskiy2020image,zhou2023nnformer} effectively capture long-range dependencies but demand extensive datasets for optimal performance. Hybrid models, like UTNet~\cite{gao2021utnet} and MedFormer~\cite{gao2022data} enhance segmentation performance by combining convolutional neural networks (CNNs) and transformers. Despite these advances, current methods still fail to meet clinical precision requirements, particularly in challenging basal and apical regions~\cite{campello2021multi}. These limitations often necessitate time-consuming manual corrections by experts, highlighting the gap between automated methods and clinical precision, especially for large-scale applications.

\begin{figure}[t]
\centering
\includegraphics[width=0.7\textwidth]{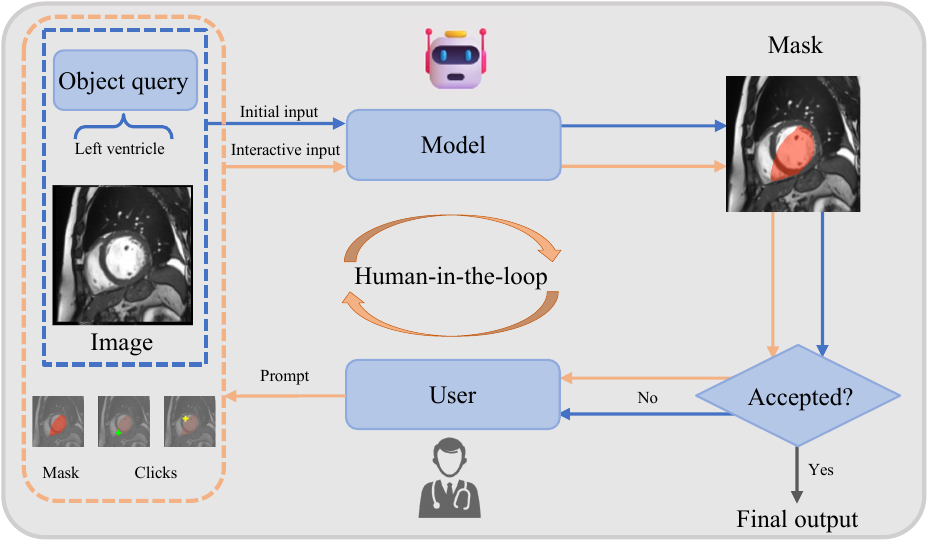} 
\caption{Illustration of the proposed versatile segmentation framework.
Our model accepts an object query, such as left ventricle, to automatically segment the target in the image. If the initial segmentation is unsatisfactory, users can refine the mask by providing corrective clicks until the final output mask can meet clinical accuracy.}
\label{figure1}
\end{figure}

To address the limitations of fully automatic segmentation approaches, interactive image segmentation methods have been proposed to facilitate efficient segmentation and annotation~\cite{liu2023simpleclick,sofiiuk2022reviving,xu2016deep}. These methods commonly rely on user clicks to guide learning-based models, mimicking human corrections and significantly improving efficiency compared to traditional pixel-by-pixel annotations. Early work~\cite{xu2016deep} first introduced CNN-based interactive segmentation, while RITM~\cite{sofiiuk2022reviving} enhanced it by integrating mask information from previous iterations. Recently, ViT-based models like SimpleClick~\cite{liu2023simpleclick} and SegNext~\cite{liu2024rethinking} have achieved state-of-the-art performance by leveraging large-scale ViT architectures and extensive pretraining on natural image datasets. However, these advancements rely heavily on advanced backbones and high-quality data, while their use of point-based prompts remains rudimentary. Notably, no interactive segmentation models have been tailored for cardiac MRIs, despite its pressing clinical demand and heavy reliance on user inputs. This gap highlights the urgent need for more efficient interactive segmentation solutions.

Recent advances in universal image segmentation~\cite{cheng2022masked,gao2024training} have explored efficient designs that combine learnable queries~\cite{ding2024clustering,adaptive_click,yan2023liver} with transformer decoders~\cite{carion2020end}. These works have inspired us to propose a novel image segmentation paradigm that leverages both data priors and human expert intelligence. 
As illustrated in Fig.~\ref{figure1}, we introduce \texttt{VerSe}, a \texttt{Ve}rsatile cardiac MRI \texttt{Se}gmentation model that integrates automatic and interactive segmentation into a unified framework. Central to our approach is the introduction of multi-query integration, which serves as prompts for a shared segmentation backbone. This design enables our model to operate seamlessly across multiple modes, enhancing its flexibility and adaptability to diverse segmentation tasks.

Our contributions are summarized as follows: (1) We introduce a novel cardiac MRI segmentation paradigm that unifies automatic and interactive segmentation, bridging the gap between learning-based methods and large-scale clinical usage through the fusion of machine and human intelligence. (2) We propose a new image segmentation architecture which can integrate multiply queries,~\textit{i.e.}, object query and click query, to prompt a foundational segmentation backbone. (3) We conducted extensive experiments on seven cardiac MRI datasets. The results demonstrate the high accuracy, efficiency, and versatility of our segmentation model. Our method also shows the potential for generalization to other out-of-distribution medical imaging segmentation tasks.

\section{Method}
\label{secmethod}

\subsection{Overview}
\begin{figure}[t]
    \centering
    \includegraphics[width=1.0\textwidth]{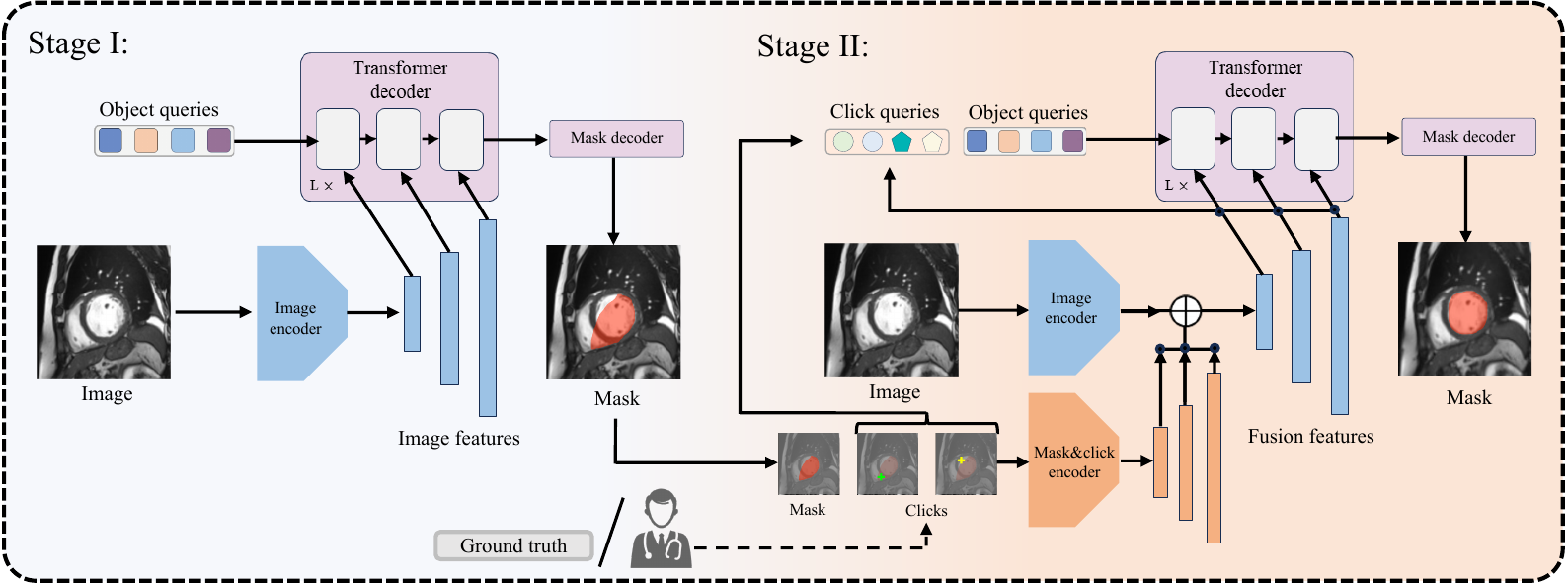} 
    \caption{Overview of the \texttt{VerSe} architecture. In stage \RNum{1}, object queries are used to automatically segment a target in the image. In stage \RNum{2}, user provides clicks as prompts to refine the initial segmentation mask. \texttt{VerSe} also supports a pure interactive mode, where the initial mask is empty and the object queries aren't activated. The image encoder, transformer decoder and mask decoder are shared across all stages. Implementation details are described in Sec.~\ref{sec:implementation}.}
    \label{figure2}
\end{figure}

We propose a novel \texttt{Ver}satile image \texttt{Se}gmentation model, \texttt{VerSe}, through the integration of object query and click query. An overview of \texttt{VerSe} architecture is shown in Fig.~\ref{figure2}. In the following, we give details of the proposed method.
\subsection{Multi-Query Integration as Prompt}
\label{secmqip}
To enable multiple functions within our model, we introduce the \textbf{Multi-Query Integration} mechanism, which guides the model to focus on relevant objects. For the automatic segmentation task, instead of using CLIP-based semantic embeddings \cite{liu2023clip}, we employ learnable object queries $\boldsymbol{X}_{o}$, as they provide stronger task-specific prior knowledge~\cite{gao2024training}. Each segmentation target is represented by a small group of learnable query vectors. Specifically, for $N$ target objects, we assign $N$ groups of learnable queries $\boldsymbol{X}_{o} = [\boldsymbol{X}_{o1}, \dots, \boldsymbol{X}_{oi},  \dots, \boldsymbol{X}_{\scriptscriptstyle oN}]$, where each group $\boldsymbol{X}_{oi} \in \mathbb{R}^{M \times C}$ consists of $M$ query vectors with $C$ channels. These queries are initialized as random parameters and optimized during training.

For the interactive segmentation task based on clicks, inspired by SAM \cite{kirillov2023segment}, we encode the click locations as sparse positional queries $\boldsymbol{X}_{s}$. For a positive click set $\boldsymbol{C}_{p} = \{(x_{1}, y_{1}), \dots, (x_{\scriptscriptstyle N_{p}}, y_{\scriptscriptstyle N_{p}})\} $, we use a Point Encoder to encode each click point to a corresponding positive positional query $\boldsymbol{X}_{sp} $, generating $N_{p} \times C $ vectors. The same process applies for the negative click set $\boldsymbol{C}_{n}$ and the corresponding negative positional queries are denoted as $\boldsymbol{X}_{sn} $.

\begin{figure*}[t]
    \centering
    \includegraphics[width=1.0\textwidth]{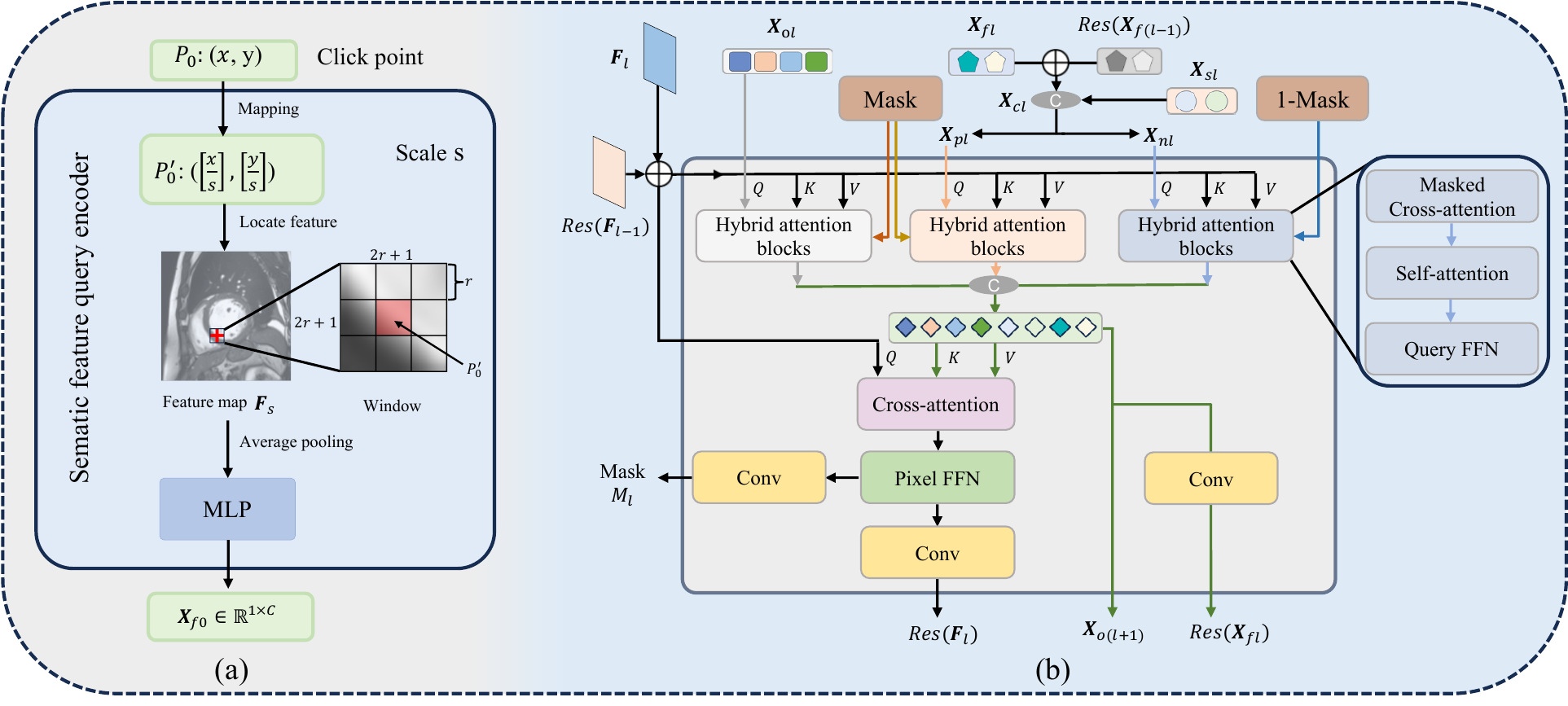} 
    \caption{(a) Process of generating semantic feature query $\boldsymbol{X}_{f0}$ for a specified click point $P_0$ at scale $s$. The original click $P_0$ is mapped to the coordinates $P_{0}^{'}$  on the down-scaling feature map. A feature patch centered at $P_{0}^{'}$ undergoes average pooling, and the resulting feature is transformed via an MLP to produce $\boldsymbol{X}_{f0}$.
    (b) Transformer decoder at the $l$-th layer. Object queries $\boldsymbol{X}_{ol}$, positive click queries $\boldsymbol{X}_{pl}$ and negative click queries $\boldsymbol{X}_{nl}$ first interact with image features $\boldsymbol{F}_{l}$, to capture foreground target and background context. These queries are then concatenated to update the image features. $\boldsymbol{X}_{fl}$, $\boldsymbol{X}_{sl}$, and $\boldsymbol{X}_{cl}$ denote semantic feature queries, sparse positional queries, and click queries at the $l$-th layer, respectively.}

    \label{figure3}
\end{figure*}

To further enhance the effectiveness of click prompts, we propose a novel \textbf{Semantic Feature Query} $\boldsymbol{X}_{f}$, as illustrated in Fig.~\ref{figure3}(a). The image encoder extracts multiple down-scaling features from the input image. For a feature map $\boldsymbol{F}_s$ with shape $(H/s, W/s)$, where $s$ is the down-scaling factor and $H, W$ are the original image height and width, we first map the original click point $ P_{0} = (x, y) $ to its corresponding coordinates in the down-scaling feature map $P_{0}^{'} =(x', y')$:


\begin{equation}    
x^{'} = \left\lfloor \frac{x}{s} \right\rfloor, \quad y^{'} = \left\lfloor \frac{y}{s} \right\rfloor.
\end{equation}
We then extract features from a $ (2r+1) \times (2r+1) $ window around $P_{0}^{'}$ and apply average pooling to compute $\boldsymbol{f}_{\text{pooled}}$:
\begin{equation}
    \boldsymbol{f}_{\text{pooled}} = \frac{1}{(2r+1)^2} \sum_{i=-r}^{r} \sum_{j=-r}^{r} \boldsymbol{F}_s(x' + i, y' + j).
\end{equation}
Finally, $\boldsymbol{f}_{\text{pooled}}$ is projected into a new feature space via a multilayer perceptron (MLP):
\begin{equation} \boldsymbol{X}_{f0} = \text{MLP}(\boldsymbol{f}_{\text{pooled}}), \end{equation}
where $\boldsymbol{X}_{f0} \in \mathbb{R} ^{1 \times C}$ represents the semantic feature query for the click at $P_{0}$. Positive and negative click sets generate semantic feature queries $\boldsymbol{X}_{fp}$ and $\boldsymbol{X}_{fn}$, respectively, at three feature scales $s=2, 4, 8$. 

$\boldsymbol{X}_{o}$, $\boldsymbol{X}_{s}$, and $\boldsymbol{X}_{f}$ collaboratively enable the versatile functions of our model. Specifically, click queries $\boldsymbol{X}_{c}$ are formed by combining sparse positional queries $\boldsymbol{X}_{s}$ and semantic feature queries $\boldsymbol{X}_{f}$. Leveraging multi-query integration, \texttt{VerSe} flexibly supports three working modes: 

\begin{itemize}
\item[$\bullet$] \textbf{Mode-1:} Automatic segmentation directed by object queries $\boldsymbol{X}_{o}$; 
\item[$\bullet$] \textbf{Mode-2:} Interactive refinement of the initial segmentation from Mode-1 by combining object queries $\boldsymbol{X}_{o}$ and click queries $\boldsymbol{X}_{c}$; 
\item[$\bullet$] \textbf{Mode-3}: Purely interactive segmentation guided by user clicks, relying solely on click queries $\boldsymbol{X}_{c}$.
\end{itemize}

\subsection{Foreground-Background Masked Attention}
\label{secfbma}

After obtaining all queries, we update the image features by attending to the multiple queries. As shown in Fig.~\ref{figure3}(b), we use foreground-background masked attention~\cite{cheng2022masked,cheng2024putting} to guide multiple queries in focusing on the target-related image features, as well as gathering background information. Unlike~\cite{cheng2024putting}, we explicitly separate the foreground target from the background using foreground and background click prompts. Specifically, we define positive click queries $\boldsymbol{X}_{p}= \left\{\boldsymbol{X}_{sp}, \boldsymbol{X}_{fp}\right\}$ and negative click queries $\boldsymbol{X}_{n}=\left\{\boldsymbol{X}_{sn}, \boldsymbol{X}_{fn}\right\}$. To ensure $\boldsymbol{X}_{p}$ and $\boldsymbol{X}_{n}$ have the same size of $N_1 \times C$, 
we add dummy points to pad $\boldsymbol{X}_{p}$ and $\boldsymbol{X}_{n}$ when click number is smaller than $N_1$. 
In our implementation, we set $N_1=24$. 
Our foreground-background masked cross-attention is computed as:
\begin{equation}
\boldsymbol{X}'_{ol} = \text{softmax}(\boldsymbol{M}_{pl} + \boldsymbol{Q}_{ol} \boldsymbol{K}_{l}^T) \boldsymbol{V}_l + \boldsymbol{X}_{ol},
\end{equation}
\begin{equation}
\boldsymbol{X}'_{pl} = \text{softmax}(\boldsymbol{M}_{pl} + \boldsymbol{Q}_{pl} \boldsymbol{K}_{l}^T) \boldsymbol{V}_l + \boldsymbol{X}_{pl},
\end{equation}
\begin{equation}
\boldsymbol{X}'_{nl} = \text{softmax}(\boldsymbol{M}_{nl} + \boldsymbol{Q}_{nl} \boldsymbol{K}_{l}^T) \boldsymbol{V}_l + \boldsymbol{X}_{nl},
\end{equation}
where $ l $ is the layer index of the transformer decoder shown in Fig.~\ref{figure2} and Fig.~\ref{figure3}(b). 
$\boldsymbol{Q}_{*l}$ is the learned linear transformation of $\boldsymbol{X}_{*l}$, where $*$ denotes different query types,~\textit{i.e.}, object, positive and negative click queries.
$\boldsymbol{K}_l, \boldsymbol{V}_l$ are the learned linear transformations of image features $\boldsymbol{F}_{l}$.
$\boldsymbol{M}_{pl}, \boldsymbol{M}_{nl} \in \{0, -\infty\}^{N_1 \times H_{l}W_{l}} $ are used to control the foreground-background attention masking. 
At the $l$-th layer, we first generate a mask prediction $\boldsymbol{M}_l$ from the pixel features $\boldsymbol{F}_{l-1}$ of the previous $(l-1)$-th layer, using a simple mask decoder followed by proper mask resizing. 
Then, $\boldsymbol{M}_{pl}$ and $\boldsymbol{M}_{nl}$ at feature location $(x,y)$ are calculated as:

\begin{equation}
\boldsymbol{M}_{pl}(x, y) = 
\begin{cases} 
      0, &  \boldsymbol{M}_l(x, y) \geq 0.5 \\
      -\infty, & \text{otherwise} 
\end{cases}, \quad
\boldsymbol{M}_{nl}(x, y) = 
\begin{cases} 
      0, &  \boldsymbol{M}_l(x, y) < 0.5 \\
      -\infty, & \text{otherwise} 
\end{cases}.
\end{equation}
Next, $\boldsymbol{X}'_{ol}, \boldsymbol{X}'_{pl},  \boldsymbol{X}'_{nl}$ are processed separately through self-attention and query feedforward networks (FFN). The resulting queries are concatenated along the query number dimension, then transformed into the \textbf{Key} and \textbf{Value} spaces using learned linear transformations. A cross-attention layer, with image features as the \textbf{Query}, followed by a pixel FFN, updates the image features, which are finally used to generate the segmentation mask.


\subsection{Residual Connection with Different Scales}
\label{secrcds}
In our multi-scale transformer decoder, the integrated queries forward through different decoder layers, continuously interacting with image features across scales. To enhance the interaction between image features and integrated queries across various scales, we adopt a residual \cite{he2016deep} resampling connection similar to the U-Net \cite{ronneberger2015u} decoder. For the image feature map  $\boldsymbol{F}_{l}$ at layer $l$, it's first resampled to align with the resolution of $\boldsymbol{F}_{l+1}$ at next layer $l+1$:

\begin{equation}
   \boldsymbol{F}_{l}^{\text{resampled}} = \text{Resample}(\boldsymbol{F}_{l}, \text{size}(\boldsymbol{F}_{l+1})).
\end{equation}
Here, \(\text{Resample}(\cdot)\) can involve upsampling or downsampling depending on the change in scale. Next, we apply a convolutional layer to calculate the residual $\text{Res}(\boldsymbol{F}_{l})$ before adding it to the feature at the next layer. The updated feature map at scale $l+1$ is computed as:
\begin{equation}
   \boldsymbol{F}_{l+1}^{\text{updated}} = \boldsymbol{F}_{l+1} + \text{Res}(\boldsymbol{F}_{l}) = \boldsymbol{F}_{l+1} + \text{Conv}(\boldsymbol{F}_{l}^{\text{resampled}}).
\end{equation}
A similar operation is applied to semantic feature queries $\boldsymbol{X}_{\text{f}}$. By iteratively resampling, computing residuals, and updating features at each scale, the model effectively integrates multi-scale context, enhancing the interaction between features and queries across scales.

\subsection{Implementation Details}
\label{sec:implementation}
\textbf{Image Encoder}. We utilize UTNet \cite{gao2021utnet} as the image encoder, which is specifically designed for cardiac MRI segmentation. 
We extract multi-scale features at resolutions of 1/8, 1/4, and 1/2 of the original image, providing a comprehensive feature representation across different scales ($S=3$).

\noindent \textbf{Mask\&Click Encoder}. 
We use simple consecutive convolutional layers to encode the dense mask\&click  feature map. Starting with original dimensions of $(3, H, W)$, the layers downsample it to $1/8$, $1/4$, and $1/2$ of the original size, matching the feature scales of the image encoder for seamless feature addition.


\noindent \textbf{Click  Representation}. During training and inference, clicks are represented as disk maps with a fixed radius of 1. Following prior work~\cite{liu2023simpleclick,sofiiuk2022reviving}, clicks are simulated by comparing current segmentation with ground truth. Unlike RITM~\cite{sofiiuk2022reviving}, we place new clicks at the center of the largest connected component in misclassified regions, better aligning with user interactions in medical imaging. In Mode-$1$ and Mode-$2$, the model generates an initial mask, using object queries, and refines it with $2$ clicks. In Mode-$3$, it iteratively uses $3$ clicks without object queries to produce $3$ segmentation masks.

\noindent \textbf{Transformer  Decoder}. We use our transformer decoders
proposed in Sec.~\ref{secmethod}
with $L = 2$ (\textit{i.e.}, $6$ layers in total). Similar to~\cite{cheng2022masked}, we use a round robin approach for $L\times$ multi-scale interaction between image features and integrated queries.

\noindent \textbf{Training Settings and Strategy}. Our models are trained for $75$ epochs with a batch size of $8$ on $8$ Quadro RTX 8000 GPUs. Images are resized to $256 \times 256$ during both training and inference. We use a combination of binary cross-entropy loss and Dice loss \cite{milletari2016v} to compute the mask loss $\mathcal{L} = \lambda_{\text{ce}} \mathcal{L}_{\text{ce}} + \lambda_{\text{dice}} \mathcal{L}_{\text{dice}}$, where $\lambda_{\text{ce}} = \lambda_{\text{dice}} = 5.0$. To enhance model robustness, we employ the following data augmentation techniques: 1) Random flips along both horizontal and vertical axes, 2) Random 90-degree rotations, 3) Random adjustments to brightness, and 4) Random adjustments to contrast. To support the diverse working modes of \texttt{Verse} during evaluation, we adopt a random training strategy. More specifically, each training batch is randomly assigned to Mode-$1$\&$2$ or Mode-$3$.

\section{Experiments}
\label{sec:experiments}
\noindent \textbf{Datasets.}  
We conducted experiments on nine publicly available datasets, including seven cardiac MRI datasets and two out-of-distribution medical imaging datasets, as summarized in Table~\ref{table_0}.
While 3D image volumes are provided in these datasets, we performed all experiments on 2D image slices, excluding slices without any instances.
Our combined cardiac training set includes three types of cardiac MRI data: 
(1) \textbf{Balanced Steady-State Free Precession (bSSFP)} sequences, which focus on segmenting the left ventricle (LV), right ventricle (RV), and myocardium wall (Myo); 
(2) \textbf{T2-weighted (T2)} sequences, which highlight myocardial edema; and 
(3) \textbf{LGE} sequences, which highlight myocardial scar. For the M\&Ms-2 dataset, we utilized only long-axis (LA) cine MRI images, to enhance data diversity, while other bSSFP datasets were acquired as short-axis images. For out-of-distribution evaluation, we tested our method on OAIZIB~\cite{ambellan2019automated}, a knee MRI dataset acquired with \textbf{Double Echo Steady State (DESS)} sequence, and BraTS~\cite{baid2021rsna}, a brain \textbf{T2-weighted} MRI dataset. These datasets allow us to assess the generalizability of our model to different domains. 

\begin{table}[h!]
    \centering
    \scriptsize
    \caption{Datasets statistics. The upper cardiac MRI datasets are for upstream training and analysis. The bottom two out-of-distribution datasets are for downstream tasks on assessing generalizability of interactive segmentation models.}
    \label{table_0}
    \begin{tabular}{lcccccc}
        \toprule
        \multirow{2}{*}{\textbf{Dataset}} & \multirow{2}{*}{\textbf{Targets}} & \multirow{2}{*}{\textbf{Modality}} & \multicolumn{2}{c}{\textbf{3D Volumes}} & \multicolumn{2}{c}{\textbf{2D Slices}}  \\
        \cmidrule(lr){4-5} \cmidrule(lr){6-7} 
        & & & \textbf{Train} & \textbf{Test} & \textbf{Train} & \textbf{Test} \\
        \midrule
        ACDC~\cite{bernard2018acdc} & LV, Myo, RV & bSSFP & 200 & 100 & 1841 & 1001 \\
        M\&Ms~\cite{campello2021multi} & LV, Myo, RV & bSSFP & 300 & 340 & 2475 & 2821 \\
        M\&Ms-2~\cite{campello2021multi} & LV, Myo, RV & bSSFP & 400 & 320 & 400 & 320 \\
        MyoPS++\cite{ding2023aligning,li2023myops,qiu2023myops,zhuang2019multivariate} & LV, Myo, RV & bSSFP & 90 & 29 & 672 & 194\\
        MyoPS++\cite{ding2023aligning,li2023myops,qiu2023myops,zhuang2019multivariate} & Myocardial Edema & T2 & 38 & 12 & 231 & 79\\
        MyoPS++\cite{ding2023aligning,li2023myops,qiu2023myops,zhuang2019multivariate} & Myocardial Scar & LGE & 56 & 18 & 313 & 98\\
        LASCARQS++~\cite{li2020atrial,li2021atrialgeneral,li2022atrialjsqnet,li2022medical} & Left Atrial  & LGE & 98 & 32 & 3561 & 1183\\
        \midrule
        OAIZIB~\cite{ambellan2019automated}& Knee Bone\&Cartilage & DESS & -- & -- & -- & 150 \\
        BraTS~\cite{baid2021rsna} & Brain Tumor & T2 & -- & -- & -- & 369\\
        \bottomrule
    \end{tabular}
\end{table}

\noindent \textbf{Baseline Methods}. For automatic image segmentation, we compared our method with nnU-Net~\cite{isensee2021nnu}, TransUNet~\cite{chen2021transunet}, UTNet~\cite{gao2021utnet}, Swin-Unet~\cite{cao2022swin} and MedFormer~\cite{gao2022data}, all of which are widely recognized and applied in medical image segmentation. For interactive image segmentation, we compared our method with RITM~\cite{sofiiuk2022reviving}, iSegformer~\cite{liu2022isegformer}, SimpleClick~\cite{liu2023simpleclick} and SegNext~\cite{liu2024rethinking}. Among these, SimpleClick and SegNext are the state-of-the-art (SOTA) methods for interactive segmentation, demonstrating exceptional performance across various benchmarks.

\noindent \textbf{Evaluation Metrics}. We use the 2D Dice score to evaluate the segmentation accuracy, which is a standard measure in medical image segmentation \cite{bernard2018acdc,gao2021utnet}.
We use the Number of Clicks (NoC) metric to assess the number of clicks required to reach a specified Dice score. Target Dice scores are set at 80\%, 85\%, 90\%, and 95\%, denoted as Noc80, NoC85, NoC90, and NoC95, respectively. Each instance allows a maximum of 20 clicks. In addition, we evaluated segmentation quality using the average Dice score of all instances at a fixed number of clicks, Dice($n$).

\begin{table}[t!]
    \centering
    \scriptsize 
    \caption{Model performance comparison for different working mode settings and different cardiac MRI datasets. Our method is the only one that can work in both automatic and interactive image segmentation modes.}
    \resizebox{\textwidth}{!}{ 
    \begin{tabular}{c|cc|cccccc}
        \toprule
         \multirow{2}{*}{\textbf{Dataset}} & \multicolumn{2}{c}{\textbf{Automatic}} & \multicolumn{5}{|c}{\textbf{Interactive}} \\
        \cmidrule(lr){2-3} \cmidrule(lr){4-9}
         &\textbf{Model} & Dice\thinspace\textuparrow & \textbf{Model} & Dice(1)\thinspace\textuparrow & Dice(20)\thinspace\textuparrow & NoC85\thinspace\textdownarrow & NoC90\thinspace\textdownarrow & NoC95\thinspace\textdownarrow \\
        \midrule
        \multirow{6}{*}{ACDC (bSSFP)} 
        & nnUNet & \textbf{89.796} & RITM & 83.999 & 92.911 & 2.175 & 4.784 & 11.055   \\
        & TranUNet & 87.649  & iSegformer & 66.595 & 92.498 & 4.251 & 7.371 & 12.384  \\
        & UTNet & 89.604 & SimpleClick & 88.050 & 92.293 & 2.402 & 5.468 & 11.647  \\
        & SwinUNet & 87.883 & SegNext & 87.665 & 92.265 & 2.295 & 5.397 & 11.794  \\
        &Medformer & 86.931 & \cellcolor[gray]{0.9} \texttt{VerSe} (Mode-3) & \cellcolor[gray]{0.9}89.757 & \cellcolor[gray]{0.9}96.698  & \cellcolor[gray]{0.9}1.485 & \cellcolor[gray]{0.9}2.352  & \cellcolor[gray]{0.9}6.480  \\
        
        &\cellcolor[gray]{0.9} \texttt{VerSe} (Mode-1) & \cellcolor[gray]{0.9}89.788 & \cellcolor[gray]{0.9} \texttt{VerSe} (Mode-2) & \cellcolor[gray]{0.9}\textbf{92.091} & \cellcolor[gray]{0.9}\textbf{96.865}  & \cellcolor[gray]{0.9}\textbf{0.431} & \cellcolor[gray]{0.9}\textbf{1.248}  & \cellcolor[gray]{0.9}\textbf{5.636}  \\
        \midrule
        \multirow{6 }{*}{M\&Ms (bSSFP)} 
        & nnUNet & 85.417 & RITM & 80.583 & 92.855 & 2.709 & 5.772 & 12.146   \\
        & TranUNet & 85.212 & iSegformer & 62.285 & 91.635 & 4.947 &  8.161  & 13.418  \\
        & UTNet & 85.998 & SimpleClick & 85.335 & 91.289 & 3.787 & 6.809 & 13.216 \\
        & SwinUNet & 84.363 & SegNext & 85.197 & 91.676 & 3.375 & 6.502  & 13.209  \\
        &Medformer & 84.472 & \cellcolor[gray]{0.9} \texttt{VerSe} (Mode-3) & \cellcolor[gray]{0.9}87.460 & \cellcolor[gray]{0.9}96.827  & \cellcolor[gray]{0.9}1.968 & \cellcolor[gray]{0.9}3.074  & \cellcolor[gray]{0.9}6.828  \\
        
        &\cellcolor[gray]{0.9} \texttt{VerSe} (Mode-1) & \cellcolor[gray]{0.9}\textbf{86.527} & \cellcolor[gray]{0.9} \texttt{VerSe} (Mode-2) & \cellcolor[gray]{0.9}\textbf{89.429} & \cellcolor[gray]{0.9}\textbf{96.887}  & \cellcolor[gray]{0.9}\textbf{1.013} & \cellcolor[gray]{0.9}\textbf{2.122}  & \cellcolor[gray]{0.9}\textbf{5.906}  \\
        \midrule
        \multirow{6}{*}{M\&Ms-2 (bSSFP)} 
        & nnUNet & \textbf{90.638} & RITM & 85.829 & 93.606 & 3.214 & 6.082 & 10.474   \\
        & TranUNet & 89.543  & iSegformer & 58.925 & 92.860 & 6.828 & 9.066 & 12.430  \\
        & UTNet & 90.266 & SimpleClick & 88.880 & 93.951 & 2.004 & 4.376  & 9.619  \\
        & SwinUNet & 89.098 & SegNext & 87.907 & 93.599 & 2.196  &  5.050 & 10.366  \\
        & Medformer & 87.544 & \cellcolor[gray]{0.9} \texttt{VerSe} (Mode-3) & \cellcolor[gray]{0.9}88.357 & \cellcolor[gray]{0.9}\textbf{97.057}  & \cellcolor[gray]{0.9}1.800 & \cellcolor[gray]{0.9}3.015  & \cellcolor[gray]{0.9}7.671  \\
        
        &\cellcolor[gray]{0.9} \texttt{VerSe} (Mode-1) &\cellcolor[gray]{0.9}86.496 & \cellcolor[gray]{0.9} \texttt{VerSe} (Mode-2) & \cellcolor[gray]{0.9}\textbf{89.830} & \cellcolor[gray]{0.9}97.013  & \cellcolor[gray]{0.9}\textbf{0.884} & \cellcolor[gray]{0.9}\textbf{2.194}  & \cellcolor[gray]{0.9}\textbf{7.047}  \\
        
        \midrule
        \multirow{6}{*}{MyoPS++ (bSSFP)} 
        & nnUNet & 88.316 & RITM & 84.800 & 92.694 & 2.311 & 5.807 & 12.635    \\
        & TranUNet & 86.886  & iSegformer & 66.061 & 92.282 & 5.619 & 8.951 & 13.488  \\
        & UTNet & 87.583 & SimpleClick & 88.030 & 92.279 & 1.868 & 4.777 & 12.184  \\
        & SwinUNet & 86.743 & SegNext & 86.431 & 92.422 & 2.189 & 5.502 & 12.551  \\
        &Medformer & 87.417 & \cellcolor[gray]{0.9} \texttt{VerSe} (Mode-3) & \cellcolor[gray]{0.9}88.834 & \cellcolor[gray]{0.9}97.079  & \cellcolor[gray]{0.9}1.577 & \cellcolor[gray]{0.9}2.624  & \cellcolor[gray]{0.9}7.047  \\

        &\cellcolor[gray]{0.9} \texttt{VerSe} (Mode-1) & \cellcolor[gray]{0.9}\textbf{88.657} & \cellcolor[gray]{0.9} \texttt{VerSe} (Mode-2) & \cellcolor[gray]{0.9}\textbf{91.071} & \cellcolor[gray]{0.9}\textbf{97.141}  & \cellcolor[gray]{0.9}\textbf{0.595} & \cellcolor[gray]{0.9}\textbf{1.526}  & \cellcolor[gray]{0.9}\textbf{6.032}  \\
                
        \midrule
        \multirow{6}{*}{MyoPS++ (T2)} 
        & nnUNet & 71.396 & RITM & 75.843 & 91.496 & 3.494 & 6.797 & 18.759   \\
        & TransUNet & \textbf{71.499} & iSegformer & 52.826 & 90.717 & 5.076 &  8.886  & 19.038  \\
        & UTNet & 70.251 & SimpleClick & \textbf{79.231} & 90.391 & \textbf{2.582} &  \textbf{6.253} & 19.481  \\
        & SwinUNet & 53.753 & SegNext & 72.492 & 91.262 & 4.101 & 7.873 & 19.342  \\
        & Medformer & 63.140 & \cellcolor[gray]{0.9} \texttt{VerSe} (Mode-3) & \cellcolor[gray]{0.9}74.096 & \cellcolor[gray]{0.9}\textbf{94.183}  & \cellcolor[gray]{0.9}5.557 & \cellcolor[gray]{0.9}7.772 & \cellcolor[gray]{0.9}\textbf{12.506}  \\

        &\cellcolor[gray]{0.9} \texttt{VerSe} (Mode-1) & \cellcolor[gray]{0.9}70.103 & \cellcolor[gray]{0.9} \texttt{VerSe} (Mode-2) & \cellcolor[gray]{0.9}74.594 & \cellcolor[gray]{0.9}93.131  & \cellcolor[gray]{0.9}6.253 & \cellcolor[gray]{0.9}8.241 & \cellcolor[gray]{0.9}12.683  \\
        
        \midrule
        \multirow{6}{*}{MyoPS++ (LGE)} 
        & nnUNet & 44.412& RITM & 60.234 & 86.735 & 7.694 & 14.531 & 19.878   \\
        & TransUNet & 55.548 & iSegformer & 38.602 & 85.691 & 12.184 & 18.306  & 20.000   \\
        & UTNet & 42.436 & SimpleClick & 65.232 & 84.773 & 8.010 & 13.541 & 20.000  \\
        & SwinUNet & 41.365 & SegNext & 61.153 & 86.879 & 9.194 & 15.367 & 19.643  \\
        & Medformer & 36.821 & \cellcolor[gray]{0.9} \texttt{VerSe} (Mode-3) & \cellcolor[gray]{0.9}\textbf{66.293} & \cellcolor[gray]{0.9}\textbf{95.024}  & \cellcolor[gray]{0.9}\textbf{6.275} & \cellcolor[gray]{0.9}\textbf{8.735}  & \cellcolor[gray]{0.9}\textbf{13.939}  \\

        &\cellcolor[gray]{0.9} \texttt{VerSe} (Mode-1) & \cellcolor[gray]{0.9}\textbf{57.879} & \cellcolor[gray]{0.9} \texttt{VerSe} (Mode-2) & \cellcolor[gray]{0.9}65.238 & \cellcolor[gray]{0.9}94.752  & \cellcolor[gray]{0.9}6.867 & \cellcolor[gray]{0.9}9.531  & \cellcolor[gray]{0.9}14.235  \\
        \midrule
        
        \multirow{6}{*}{LAScarQS++ (LGE)} 
        & nnUNet & 84.827 & RITM & 82.957 & 92.704 & 3.550 & 6.163 & 13.866   \\
        & TranUNet & 83.441  & iSegformer & 72.017 & 92.750 & 4.527 & 7.546 & 14.896  \\
        & UTNet & \textbf{85.275} & SimpleClick & 84.056 & 91.279 & 3.724 &  6.433 & 14.908  \\
        & SwinUNet & 79.786 & SegNext & 84.284 & 94.011 & 2.558 & 4.673  & 10.961  \\
        & Medformer & 82.378 & \cellcolor[gray]{0.9} \texttt{VerSe} (Mode-3) & \cellcolor[gray]{0.9}85.841 & \cellcolor[gray]{0.9}\textbf{97.452}  & \cellcolor[gray]{0.9}2.103 & \cellcolor[gray]{0.9}2.940  & \cellcolor[gray]{0.9}5.717 
        
        \\
        &\cellcolor[gray]{0.9} \texttt{VerSe} (Mode-1) & \cellcolor[gray]{0.9}83.159 & \cellcolor[gray]{0.9} \texttt{VerSe} (Mode-2) & \cellcolor[gray]{0.9}\textbf{86.919} & \cellcolor[gray]{0.9}97.172  & \cellcolor[gray]{0.9}\textbf{1.414} & \cellcolor[gray]{0.9}\textbf{2.270}  & \cellcolor[gray]{0.9}\textbf{5.077}  \\
        \bottomrule
    \end{tabular}
    }
    \label{table_1}
\end{table}
\subsection{Results}
\subsubsection{Comparison with Automatic Models.}
Table~\ref{table_1} compares our proposed \texttt{VerSe} framework with several specialized segmentation models designed for specific tasks. Despite lacking any specialized design to enhance automatic segmentation, \texttt{VerSe} achieves competitive performance on large-scale datasets such as ACDC and M\&Ms, demonstrating its robustness and adaptability across diverse segmentation scenarios. However, on more challenging datasets like MyoPS++ (T2), where all models struggle to meet clinical requirements, due to the dataset's inherent complexity, the interactive capabilities of \texttt{VerSe} become particularly advantageous. By enabling efficient user-driven refinements through click-based interactions, \texttt{VerSe} provides a practical solution to enhance segmentation accuracy in challenging cases, bridging the gap toward clinical applicability.

\subsubsection{Comparison with Interactive Models.}
\begin{figure*}[t]
    \centering
    \includegraphics[width=1.0\textwidth]{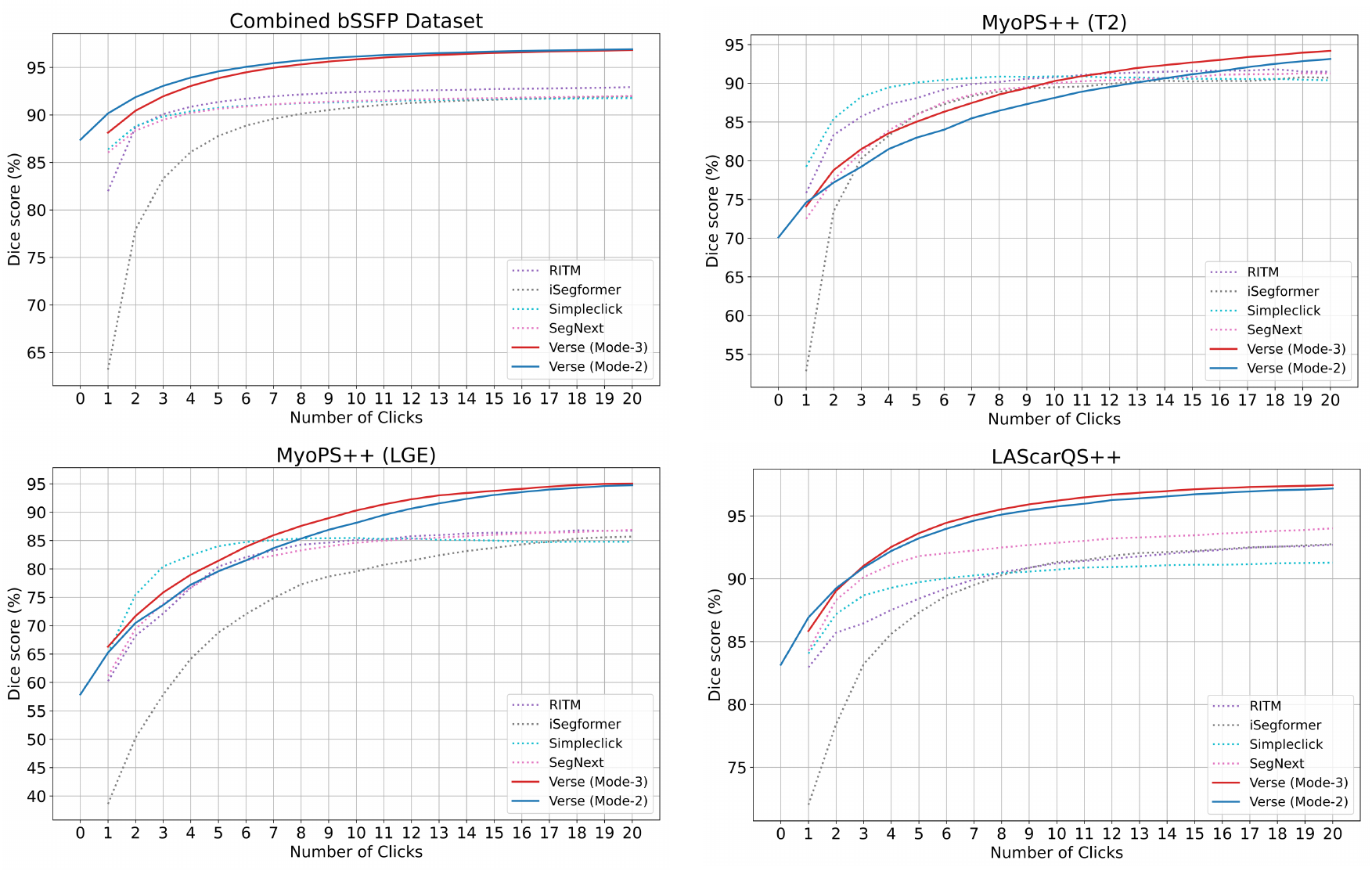} 
    \caption{Convergence analysis for models tested on four types of segmentation targets. The Combined bSSFP Dataset, including ACDC, M\&Ms, M\&Ms-2, and MyoPS++ (bSSFP), focuses on LV, Myo, and RV structures. \texttt{VerSe} demonstrates consistent accuracy improvements across all tasks as the number of clicks increases.}
    \label{figure4}
\end{figure*}

Table~\ref{table_1} highlights the performance comparison between \texttt{VerSe} (Mode-3) and previous SOTA interactive models. \texttt{VerSe} (Mode-3) consistently achieves the best Dice scores and lower interaction costs among six out of seven datasets. Notably, on larger datasets, such as ACDC, M\&Ms, and LAScarQS++, \texttt{VerSe} achieves the best performance across all metrics, significantly outperforming existing methods. For example, on the M\&Ms dataset, which has the largest number of instances, \texttt{VerSe} achieves a Dice(1) score of 89.757\%, significantly surpassing SimpleClick (85.335\%) and SegNext (85.197\%). This demonstrates VerSe's ability to handle complex and large-scale data efficiently. Furthermore, as illustrated in Fig.~\ref{figure4}, \texttt{VerSe} not only achieves faster convergence but also demonstrates steady improvements in segmentation accuracy as the number of clicks increases, highlighting the efficiency of its interactive prompting mechanism.

On smaller and more challenging datasets, such as MyoPS++ (LGE) and MyoPS++ (T2), \texttt{VerSe} continues to show significant advantages. On the MyoPS++ (LGE) dataset,
\texttt{VerSe} achieves a Dice(20) of 95.024\%, far exceeding the $\sim85$\% average of competing methods. Similarly, on the MyoPS++ (T2) dataset, while SimpleClick initially leads in the first 10 clicks, \texttt{VerSe} demonstrates more sustained improvements, ultimately achieving the best Dice(20) of 94.183\% and a NoC95 of 12.506. These results highlight \texttt{VerSe}'s ability to efficiently utilize user interactions to refine segmentation results, even in difficult scenarios.

Overall, VerSe (Mode-3) sets a new benchmark in interactive segmentation, by achieving SOTA performance with minimal interaction costs. Its robust performance across both large-scale datasets and complex segmentation tasks underscores its adaptability and effectiveness in diverse cardiac MRI applications.

\subsubsection{\texttt{VerSe} (Mode-2) vs. \texttt{VerSe} (Mode-3).}

Table~\ref{table_1} presents a detailed comparison between \texttt{VerSe} operating in Mode-2 and Mode-3 across various datasets. \texttt{VerSe} (Mode-2) achieves superior performance in 4 out of 7 datasets compared to \texttt{VerSe} (Mode-3), particularly in the ACDC and M\&Ms datasets, where the initial automatic segmentation effectively reduces user interaction while improving accuracy. In contrast, \texttt{VerSe} (Mode-3), which relies solely on interactive segmentation without automatic initialization, excels on datasets where precise automatic initialization is particularly challenging, such as MyoPS++ (T2) and MyoPS++ (LGE). These results underscore the robust interactive segmentation performance of \texttt{VerSe} (Mode-3), especially in scenarios lacking reliable automatic initialization. Meanwhile, Mode-2’s dependence on object queries highlights the need for larger and more diverse datasets to unlock its full potential. For datasets with limited training samples, Mode-3 serves as a reliable fallback. These results highlight the flexibility and adaptability of \texttt{VerSe}’s unified segmentation framework in addressing varying clinical and data-specific needs.

\subsubsection{Out-of-Distribution Evaluation.}
We trained all interactive segmentation models using cardiac MRI datasets, while we evaluated their performance on out-of-distribution datasets.
The results are summarized in Table~\ref{table_2}. On the BraTS dataset, \texttt{Verse} achieves a remarkable Dice score of 94.492\% with just 10 clicks, significantly outperforming other models. In addition, it delivers the highest Dice(20) score of 96.811\% and demonstrates the best annotation efficiency. On the OAIZIB dataset, \texttt{Verse} achieves the lowest NoC80 value of 11.787, highlighting its superior efficiency. These results collectively showcase the robust generalization capabilities of \texttt{Verse} across diverse medical imaging domains.

\begin{figure*}
    \centering
    \includegraphics[width=1.0\textwidth]{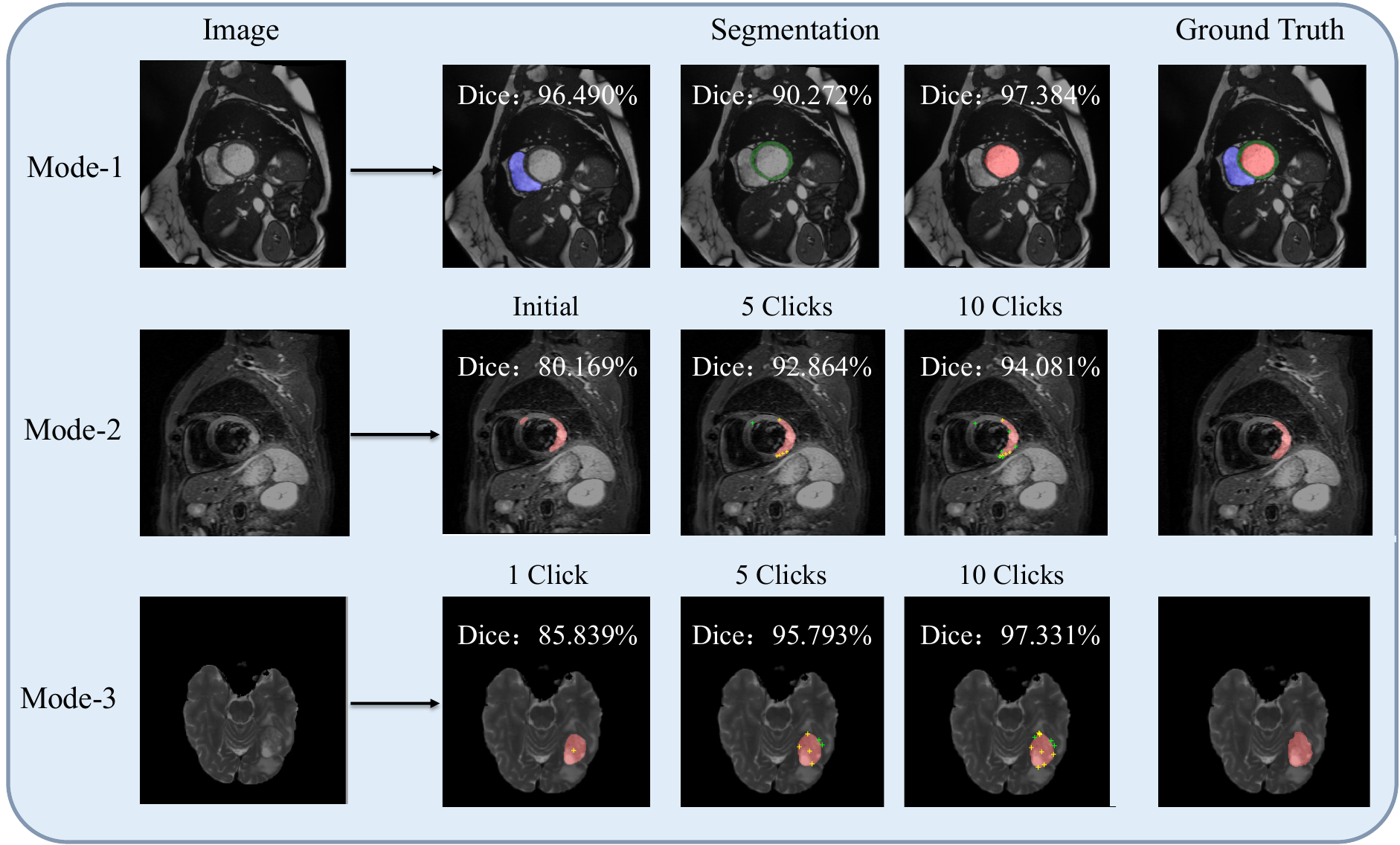} 
    \caption{Segmentation results of \texttt{VerSe} on different medical image segmentation tasks. First row: Automatic segmentation of three structures on cardiac cine MRI. Second row: Interactive refinement of myocardial edema segmentation on cardiac T2-weighted MRI. Third row: Interactive segmentation of a tumor on out-of-distribution brain MRI.}
    \label{figure5}
\end{figure*}

\subsubsection{Visualization.} In Fig.~\ref{figure5}, we demonstrate representative segmentation results of \texttt{VerSe} on cardiac MRI and out-of-distribution brain MRI. As can be seen from the results, our design of \texttt{VerSe} allows it to handle different medical image segmentation tasks with high versatility.

\begin{table}[t!]
    \centering
    \scriptsize 
    \caption{Out-of-distribution evaluation on BraTS~\cite{baid2021rsna}  and OAIZIB~\cite{ambellan2019automated} datasets. \texttt{Verse} shows strong generalizability in its interactive segmentation.}
    \begin{tabular}{c|ccc|ccc}
        \toprule
         \multirow{2}{*}{\textbf{Model}}& \multicolumn{3}{c}{\textbf{OAIZIB}} & \multicolumn{3}{|c}{\textbf{BraTS}} \\
        \cmidrule(lr){2-4} \cmidrule(lr){5-7}
         & Dice(10)\thinspace\textuparrow & Dice(20)\thinspace\textuparrow & NoC80\thinspace\textdownarrow & Dice(10)\thinspace\textuparrow & Dice(20)\thinspace\textuparrow & NoC80\thinspace\textdownarrow \\
        \midrule
        RITM & 70.334 & 79.071 & 13.988 & 88.925 & 93.898 & 5.136   \\
        iSegformer & 65.509 & 76.784  & 17.063& 86.465 & 92.281 & 6.482  \\
        SimpleClick & \textbf{73.260} & \textbf{82.523} & 12.976 & 92.537 & 93.927 & 3.436   \\
        SegNext & 57.211 & 76.261 & 17.427 & 77.448 & 91.619 & 9.604   \\
        \cellcolor[gray]{0.9} \texttt{VerSe} (Mode-3) & \cellcolor[gray]{0.9} 69.516 & \cellcolor[gray]{0.9} 80.631 & \cellcolor[gray]{0.9} \textbf{11.787} & \cellcolor[gray]{0.9} \textbf{94.492}  & \cellcolor[gray]{0.9} \textbf{96.811} & \cellcolor[gray]{0.9} \textbf{2.777} \\
        \bottomrule
    \end{tabular}
    \label{table_2}
\end{table}

\begin{table}[h!]
    \centering
    \caption{Ablation study of each model component in \texttt{VerSe} under working Mode-$3$. `S': Semantic feature queries. `F-B': Foreground-background masked attention of click queries. `R': Residual connection.}
    \resizebox{0.6\textwidth}{!}{
    \begin{tabular}{lccccccc}
        \toprule
        \textbf{Model} & \textbf{S} & \textbf{F-B} & \textbf{R} &\textbf{Dice(1)\thinspace\textuparrow} & \textbf{Dice(20)\thinspace\textuparrow} & \textbf{NoC90\thinspace\textdownarrow} & \textbf{NoC95\thinspace\textdownarrow} \\
        \midrule
        A1 &\xmark & \checkmark &\checkmark & 89.563 & 
        96.537 & 2.543 & 7.015 \\
        A2 &\checkmark & \xmark &\checkmark & 90.351 & 96.540 & 2.112 & 6.511 \\
        A3 &\checkmark & \checkmark &\xmark & 88.480 & 96.172 & 2.755 & 7.513 \\
        
        \midrule
        \texttt{Verse} &\checkmark & \checkmark &\checkmark & \textbf{91.025} & \textbf{96.659} & \textbf{2.028} & \textbf{6.372} \\
        \bottomrule
    \end{tabular}}
    \label{table_3}
\end{table}

\subsection{Ablation Studies}
In this section, we conducted ablation studies to demonstrate the effectiveness of our model design. 
All of our ablation studies were conducted on the ACDC dataset with \texttt{VerSe} working in Mode-$3$ only. The results are shown in Table~\ref{table_3}. 
(1) In model A1, we eliminated the semantic feature queries $\boldsymbol{X}_{f}$ from click queries $\boldsymbol{X}_{c}$.
(2) In model A2, we eliminated the masked attention branch (see the rightmost hybrid attention blocks in Fig.~\ref{figure3}(b)) of negative click queries but combined positive and negative click query masked attention branches as a single one.
(3) In model A3, we eliminated the residual connections across different scales.
The results show that each of the three model components are necessary for the high performance of our model.

\section{Conclusion}

In this work, we propose \texttt{VerSe}, a novel framework that unifies automatic and interactive segmentation modes, through a multi-query integration mechanism. By effectively leveraging both object and click queries, \texttt{VerSe} achieves state-of-the-art performance in both segmentation accuracy and interaction efficiency. Our experiments on seven cardiac MRI datasets and two out-of-distribution medical imaging datasets demonstrate the robustness, efficiency, and generalizability of the proposed method. \texttt{VerSe} not only bridges the gap between automatic and interactive segmentation but also sets a new benchmark for versatile image segmentation tasks. Future directions include extending the framework to handle natural data, improving scalability for larger multi-modal datasets, and enhancing interpretability for clinical adoption.

%

\bibliographystyle{splncs04}
\bibliography{mybibliography.bib}

\end{document}